# Model to Model: Understanding the Venus Flytrap Snapping Mechanism and Transferring it to a 3D-printed Bistable Soft Robotic Demonstrator

Maartje H. M. Wermelink[✉1,2[0009-0000-2337-6340]],
Renate Sachse[3,4[0000-0001-8895-9635]], Sebastian Kruppert[1,2[0000-0001-5932-8455]],
Thomas Speck[1,2[0000-0002-2245-2636]], Falk J. Tauber[1,2[0000-0001-7225-1472]]

[1] Cluster of Excellence *liv*MatS @ FIT – Freiburg Center for Interactive Materials and Bioinspired Technologies, Germany
[2] Plant Biomechanics Group (PBG) Freiburg @ Botanic Garden of the University of Freiburg, Freiburg University, Germany
[3] Chair of Structural Analysis, Department of Civil and Environmental Engineering, TUM School of Engineering and Design, Technical University of Munich, Germany
[4] John A. Paulson School of Engineering and Applied Sciences, Harvard University, Boston, USA
`maartje.wermelink@biologie.uni-freiburg.de`

**Abstract.** The Venus flytrap (*Dionaea muscipula*) does not only serve as the textbook model for a carnivorous plant, but also has long intrigued both botanists and engineers with its rapidly closing leaf trap. The trap's closure is triggered by two consecutive touches of a potential prey, after which the lobes rapidly switch from their concave open-state to their convex close-state and catch the prey within 100-500 ms after being triggered. This transformation from concave to convex is initiated by changes in turgor pressure and the release of stored elastic energy from prestresses in the concave state, which accelerate this movement, leading to inversion of the lobes' bi-axial curvature. Possessing two low-energy states, the leaves can be characterized as bistable systems. With our research, we seek to deepen the understanding of Venus flytrap's motion mechanics and apply its principles to the design of an artificial bistable lobe actuator. We identified geometrical characteristics, such as dimensional ratios and the thickness gradient in the lobe, and transferred these to two 3D-printed bistable actuator models. One actuator parallels the simulated geometry of a Venus flytrap leaf, the other is a lobe model designed with CAD. Both models display concave-convex bi-stability and snap close. These demonstrators are the first step in the development of an artificial Venus flytrap that mimics the mechanical behavior of the biological model and can be used as a soft fast gripper.

**Keywords:** Artificial Venus Flytrap, Biomimetics, Finite Element Simulation, Venus Flytrap Mechanics

# 1      Introduction

Nature has consistently served as a source of inspiration for human innovation, with the plant kingdom offering a remarkable variety of mechanisms for investigation. While most plants appear stationary at first glance, some exhibit movements at remarkable speed. Examples range from the leaf folding of *Mimosa pudica*, which is directly observable by naked eye, to the underwater prey capture of the bladderwort plant (*Utricularia australis*), which takes place in 0.5 ms and therefore is much too fast for the human eye. On land, the Venus flytrap (*Dionaea muscipula*) stands out in terms of speed. It is a remarkable example of rapid plant movement, displaying a distinctive snapping mechanism for prey capture. The trap consists of two doubly curved lobes that form bistable structures connected via a midrib. The lobes are characterized by a higher thickness in the region near the midrib and are thinner towards the edges [1]. This architecture creates a bistable system, enabling the trap to switch rapidly from the open to the closed state. When prey enters the trap and touches two trigger hairs (or the same one twice) within ~20 seconds, it initiates a cascade of events leading to trap closure in approximately 100 to 500 ms [2, 3]. This fast motion is crucial for successful prey capture, its speed and effectiveness have intrigued botanists and engineers alike, prompting further research into the mechanisms behind this rapid closure [4–9]. At the core of the Venus flytrap's rapid movement lies a complex interplay of structural properties. The transition between the stable states is initiated by a series of physiological events involving action potentials and turgor changes and are triggered by the sensitive trigger hairs on the inner surface of the trap [2, 8, 10]. Recent studies have revealed more details of this process, elaborating on how the plant harnesses stored elastic energy to achieve its speed. In an open state, the trap lobes are in a state of prestress. Upon activation, an energy barrier is overcome and this prestress is released, resulting in a curvature change of the lobe [1, 3]. Understanding the mechanics of the Venus flytrap's snapping motion offers valuable insights into the principles of bistable systems, a concept with a wide range of applications in engineering and materials science [11]. It is particularly intriguing from an engineering perspective and offers potential for energy-efficient technologies. The Venus flytrap has long been a source of inspiration in the field of robotics and soft machines, influencing designs in both rigid and soft domains [12–18], especially as role models for gripper systems [12-15, 19, 20]. However, most of these systems only transfer some characteristics of the Venus flytrap's trap, for example missing the characteristic snap buckling of the lobe and instead closing with a continuous motion [15-18, 19, 20]. Some systems utilize pneumatic actuation to drive a fast closure, necessitating a permanent and often bulky gas supply [13, 15, 17]. Two AVF gripper systems incorporate the characteristic snap [12, 19], but lack the bi-axial bending in either concave or convex state. So far there is one bistable snapping gripper with bi-axial bending [14], which is driven by dielectric elastomer actuators necessitating high voltage and power supplies [14, 20]. As such current systems often rely on external for energy supplies and few designs have successfully implemented the reversible bistability exhibited by the plant's lobes, as well as their tapered design, with many artificial Venus flytrap systems either working one-way, or exhibiting a continuous motion instead of snap-buckling.

Our research aims to bridge this gap by closely reviewing and examining the kinematics of the trap lobes during closure. We identify key features that contribute to their speed and efficiency to design an artificial Venus flytrap lobe that transfers the plant's bistable nature into a biomimetic actuator applicable for future soft robotic grippers. As a first step, we developed different bistable demonstrator lobes based on the Venus flytrap and analyzed them in terms of structural and mechanical factors contributing to the rapid closure. This approach has the potential to yield significant advancements in the field of soft robotics, where developing fast, energy-efficient gripping mechanisms remains a challenge.

## 2     Insights into the Venus Flytrap Snapping Mechanism by further studies

Previous studies have advanced the understanding of the biomechanics of the Venus flytrap, providing insights into both its rapid closure and slower reopening processes [1–6, 8, 21]. We review these insights and use them to abstract the biological trap and identify the main characteristics of the snapping mechanism, as well as the mechanics of the closing and reopening. These characteristics are foundational for the development of an artificial Venus flytrap system, and we based the design of our bio-inspired artificial lobe on these characteristics.

### 2.1     Biomechanics of the Closing Mechanism

The characteristic snap-through behavior that the Venus flytrap displays in its trap lobes, results in two stable states—open and closed—which can be maintained without continuous energy input, a hallmark of bistable structures. Forterre et al. [2] were the first to demonstrate that the rapid closure relies on a combination of active hydraulic actuation and the passive snap-buckling of the trap lobes. Sachse et al. [1] investigated the snapping of the Venus flytrap through experiments and computer simulations. Computer simulations were used to study effects and verify hypotheses regarding the multilayered lobe structure and mechanics of the closing and reopening mechanisms. They explored different strain distributions across the surface of the lobe, which are a measure for how much the motor cells expand or contract between the open and closed state. The experimental measurements are presented with digital image correlation in Fig. 1. Fig. 1A shows the measured strain of the outer epidermis in y direction. Fig. 1B maps the strain over the lobe surface during closure, determined by Sachse et al. [1] with finite element simulation models.

The geometry of the model resembled the geometry of an open trap. In addition, they investigated the contributions of different tissue layers to the actuation (two-layer and three-layer models with varying actuation). As shown in Fig. 1C, the characteristic snapping motion that the biological trap performs was not observed initially. However, by modeling a further opened and more doubly curved trap lobe geometry, they were able to include a prestress effect in the model and reproduce the rapid snap-through behavior observed in nature, illustrated in Fig. 1D. The different colored lines in the figure represent various actuation scenarios for the inner and outer epidermis, as detailed in the legend. The three-layered model, represented by the solid lines, most closely resembled the natural trap's behavior. Specifically, the model with a neutral middle layer (acting as a lever arm) and a contraction of the lobe inside by 20% of the value of the expansion of the outer lobe showed the best match to observed lobe deformation, and the experimental measurements, as illustrated in Fig. 1B and D.

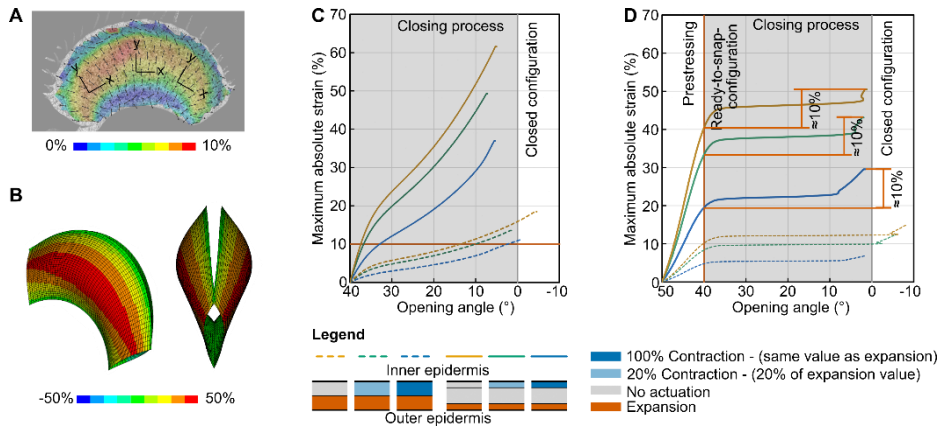

**Fig. 1.** Biomechanical analysis of Venus flytrap closure. (**A**) Digital image correlation (DIC) measurements showing the major strain distribution in the y-direction on the outer epidermis of a Venus flytrap lobe during closure. The color scale indicates strain magnitude from 0% (blue) to 10% (red). (**B**) Finite element simulation models illustrating strain distribution of the outer lobe (left) and deformed trap geometry (right) during closure. The color scale ranges from -50% (blue) to 50% (red) strain. (**C**) Closing process without prestress, showing the relationship between maximum absolute strain and opening angle for different actuation scenarios. (**D**) Closing process with prestress included, demonstrating the "ready-to-snap" configuration and subsequent closure. In both C and D, different line styles and colors represent various combinations of expansion and contraction in the outer and inner epidermis, as detailed in the legend. Solid lines represent three-layered models, which most closely resemble natural trap behavior. Parts of the figure are adapted with permission from [1] (Copyright (2020) National Academy of Sciences).

These findings led them to conclude that the trap in nature is prestressed, which was verified through dehydration experiments [1]. The trap is held in a "ready-to-snap" configuration, this prestressed state ensures fast snap closure, which is crucial for a successful prey capture. The prestress effectively reduces the energy barrier that needs to be overcome to initiate closure, allowing the trap to respond more quickly to stimuli [1]. It further underlines the conclusion of Forterre et al. [2] that the combination of active hydraulic actuation and the passive snap-buckling of the trap lobes allows the plant to overcome the limitations of its hydraulic system, achieving closure speeds that would be impossible through cellular water movement alone.

## 2.2   Biomechanics of the Reopening Mechanism

In addition to studies on the closing mechanism, Durak et al. [21] investigated the reopening process of the Venus flytrap. This study revealed that the reopening mechanism differs significantly from the closing mechanism, exhibiting more varying behaviors. While closure occurs rapidly, reopening is a much slower process taking hours, after failed capture, up to several days, after successful prey capture. Durak et al. [21] investigated the closure of different sized traps in their experiments. The median reopening time for normal-sized traps was about 28 hours, while larger traps took about 31 hours. Two distinct reopening modes were observed, depending on trap size and morphology. Smaller traps (up to 3 cm in length) generally reopen via smooth, continuous outward bending. In contrast, larger traps (up to 4.5 cm in length) could exhibit either smooth reopening or a two-phase process: slow initial bending followed by a faster, albeit small, snap-through opening. With 3D digital image correlation, it could be shown that the reopening involves complex strain distributions on both inner and outer trap surfaces. These distributions differ from those observed during closure, indicating that reopening is not simply a reversal of the closing process. Furthermore, their studies indicated that the type of opening process (snapping or continuous) depends on the slenderness (ratio of length to thickness) of the trap. More slender traps are more likely to exhibit snap-through during reopening [21].

## 3   Artificial Venus Flytrap Actuators

### 3.1   Biological Characteristics as Inspiration for a Soft Gripper

The Venus flytrap's unique biomechanics offer several advantages that make it an excellent model for soft gripper design. In reviewing the studies on the Venus flytrap, we identified the following characteristics of the Venus flytrap's movements that are critical for the translation of the motion into a gripper:

- Prestress is essential for the characteristic fast snap-closure motion.
- During snapping, the elongation or contraction of the motor cells within the different tissue layers of the Venus flytrap lead to a strain distribution over the entire trap lobe. As a result, active and inactive layer structures that contribute to the snapping mechanism were identified. This leads to the following assumptions:
  - Maximum actuation occurs in the middle of the trap, not at the edges.
  - The middle layer acts as a passive lever structure between the active outer and inner layers.
- The reopening mechanism significantly differs from the closing mechanism. Reopening can occur by either smooth bending or a two-phase process with snap-through, influenced by the slenderness of the trap.

By transferring the above-listed key characteristics, we can create grippers with enhanced capabilities and efficiency, such as rapid actuation for quick object capture, an energy-efficient bistable mechanism for fast secure holding combined with controlled reopening for careful object release. We selected and transferred several key characteristics to our first design of an artificial Venus flytrap lobe. First, we implemented a double-curved structure, which allowed for the transfer of the bi-stability of the natural trap. By this, we were able to mimic the differentiated actuation between loading and snapping observed in the natural trap. Second, we introduced a thickness taper between the hinge- and edge side as seen in the biological model. Third, we translated the geometrical proportions of the Venus flytrap lobe to 3D printed actuators. Given the fact that our actuators are single material that neither contract or expand, they represent the Venus flytrap lobe's middle layer rather than the complete lobe.

### 3.2   Model Designs

We used two approaches to design the artificial lobes, each having a version with constant and tapered thickness. All actuators were prepared as digital models and exported to STL files using FreeCAD (version 0.21.2, FreeCAD Project Association, Brussels, Belgium). The files were then sliced in PrusaSlicer (version 2.8.1+win64, Prusa Research, Prague, Czech Republic), both with the concave side facing down using support structures. They were printed on a Prusa MK4 Input Shaper FDM printer (Prusa Research, Prague, Czech Republic) with flexible TPE filament (Fiberology Fiberflex 40D, Fiberlab S.A., Brzezie, Poland). The material properties of the filament [22] and the used print settings are shown in Table 1. The lobes were printed in the 'snapped' state so by mechanically loading the lobe, it becomes prestressed and set into a ready to snap state. This is the similarity with the biological model that we aim for in this early stage of the gripper development. As described in Section 3.3, we characterized the actuators and compared them based on different factors.

**Table 1.** Material properties and print settings used in PrusaSlicer 2.8.1 [22]

| Material properties | | | Print settings | | |
|---|---|---|---|---|---|
| Physical property | Value | Unit | Parameter | Value | Unit |
| Density | 1.16 | g cm$^{-3}$ | Nozzle diameter | 0.6 | mm |
| Stress at    5% strain | 2 | MPa | Printing temperature | 220 | °C |
|              10% strain | 4 | MPa | Bed temperature | 60 | °C |
|              50% strain | 9 | MPa | Tool fans | min 20 | % |
|              break | 28 | MPa |  | max 60 | % |
| Elongation at break | 700 | % | Layer height | 0.2 | mm |
| Tear strength | 115 | kN m$^{-1}$ | Perimeters | 3 |  |
| Melting temperature | 160 | °C | Extrusion multiplier | 1.1 |  |
|  |  |  | Printing speed | 30 | mm s$^{-1}$ |
|  |  |  | Supports | On |  |
|  |  |  | Ironing | On |  |

**Simplified Geometry (SG).** One approach was to redesign the geometry of the Venus flytrap lobe, to mimic the biological curvature. We utilized a simplified version of the simulated curvature of an open Venus flytrap trap lobe that Sachse et al. [1] used for their analysis in section 2. It was imported into FreeCAD to model it as a curved structure (blue in Fig. 2A, B, C). Since the biological lobe is restrained at the hinge, we added a restraining surface at the same edge in the model design. Finally, we added a flat area for handling purposes (both orange in Fig. 2A, B, C).

**Abstracted Trap Lobe (ATL).** The other approach included abstraction of the Venus flytrap lobe morphology. The actuator was designed in FreeCAD, following the Venus flytrap trap lobe contour. To estimate the geometry for the model, we assumed the curvature of the biological lobe to fit an ellipsoid surface curve and projected the designed contour onto the surface of an ellipsoid to achieve a double-curved geometry (Fig. 2D, E). We identified a set of parameters for the geometry that was stable in both configurations. The lower edge of the model was stabilized by a rib, depicted in Fig. 2F and G.

**Thickness gradient.** Both the geometry and the lobe models were printed in two versions. One with a constant thickness of 0.93 mm, the other version with a tapered thickness, being 0.90 mm at the tip and 1.30 mm at the base. The preliminary models had thicknesses in proportion to the lobe thickness of the biological model plant, but these demonstrators were not bistable, due to the material properties of our filament. In preliminary testing, we identified the dimensions that showed the desired bi-stability. Due to the concave shape of the models and the resolution of the printer in the z-axis, the actual thickness showed a variation, which was observed to be of about 0.05 mm.

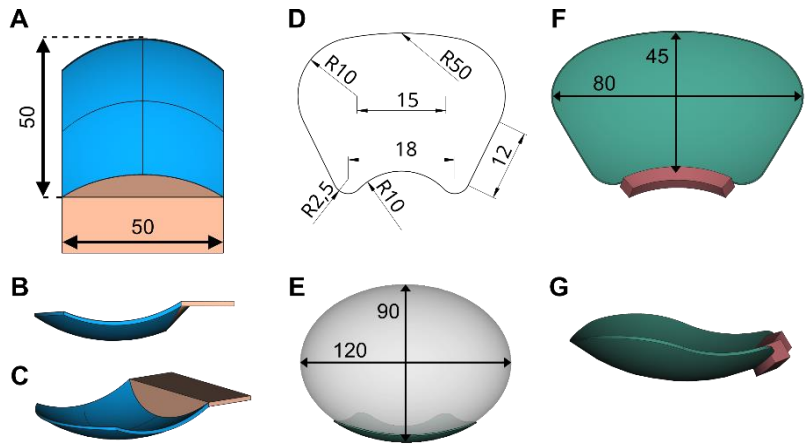

**Fig. 2.** Models with constant thickness as designed in FreeCAD (**A–C**, **F**, **G**). The SG model is shown facing the concave surface, with curves from the simplified geometrical model shown on the surface (**A**), from a side view (**B**) and with an orthographic view (**C**). The ATL model started as a flat sketch (**D**), which was projected onto an ellipsoid (**E**, projected lobe below) to create a curved structure. **F** shows the ATL facing the concave surface, and **G** from a side view. The measurements are in millimeters and the radii in degrees.

### 3.3 Model Characterization

We characterized the SG and ATL models and their two respective thickness variants with a universal testing machine Inspekt table 5, using a 50N load cell, and recorded the data in LabMaster version 2.8.13.9 (both Hegewald & Peschke Meß-und Prüftechnik GmbH, Nossen, Germany). We tested for the energy and maximum force required to change configuration from the concave (initial) state to convex state (this direction we called "loading") and vice versa ("snapping"). Table 2 shows an overview of the performed tests. For visualization purposes, photos and videos were taken with an iPhone 12 mini (both the normal (60 fps) and slow motion (120 fps, 12% speed) video modes were used). The model rested on a holder, providing the freedom of movement required for buckling while keeping it in place for the experiment. An indenter with a widened rounded end was used to enact a downwards force on the model. Holder and indenter used in the setup were designed and printed for this experiment (Fig. 3A, B).

The point of indentation was exactly on the apex of the model. Except for the loading measurements of the tapered SG model, we initially indented on the apex, but no configuration change was achieved. We needed to adjust the indentation point 3 mm away from the apex towards the hinge-side edge to achieve a configurational change and to "load" the tapered SG model. In all snapping experiments over all models the indentation point was the same, triggering the snap closure. The models were tested by indenting from above at a constant speed of 40 mm/s, until they changed configuration. Force (in N) and stroke (in mm) were recorded as time factors.

**Table 2.** Overview of the test setups, including four samples in loading and snapping direction

| Model | Thickness (mm) | Direction | Sample size (n) |
|---|---|---|---|
| SG | 0.9 | Loading | 16 |
|  |  | Snapping | 15 |
|  | 0.9–1.3 | Loading | 15 |
|  |  | Snapping | 15 |
| ATL | 0.9 | Loading | 16 |
|  |  | Snapping | 15 |
|  | 0.9–1.3 | Loading | 15 |
|  |  | Snapping | 16 |

### 3.4  Data Analysis

For statistical analysis of the data, "R" version 4.4.1 (R Development Core Team 2020) and "RStudio" version 2024.12.1.563 (RStudio Team 2020) were used [23, 24]. Using the Shapiro test, the data were tested for normal distribution and subsequent multi-factorial ANOVAs were conducted in combination with Tukey HSD post-hoc tests to identify significant differences between the actuators regarding work and maximum force necessary to load or snap.

## 4     Results

Eight different setups were tested, over four different actuator models. Each model was tested via indentation for the maximum force and work necessary to change from one stable state to the other. This was done for both loading of the trap and for snapping back to the original initial shape (Fig. 3C–J). The starting and buckling points could be determined from the measurements (Fig. 3K, starting point and buckling point respectively at X1 and X2). The work required was calculated from the integral of the stroke-force measurements from the start of deformation (X1) until the moment that the model starts accelerating in its transition to the other state (X2).

The change in configuration typically started at the point of indentation, which in the ATL models for both the loading and snapping tests would first spread towards the hinge-side edge, then laterally to one side, from where it propagated to the other side until the lobe was fully reversed (Fig. 3H). The SG models buckled first from the apex to the upper edge during loading, followed by the hinge-side edge. For both the ATL and the SG models, the buckling was much more abrupt than the loading.

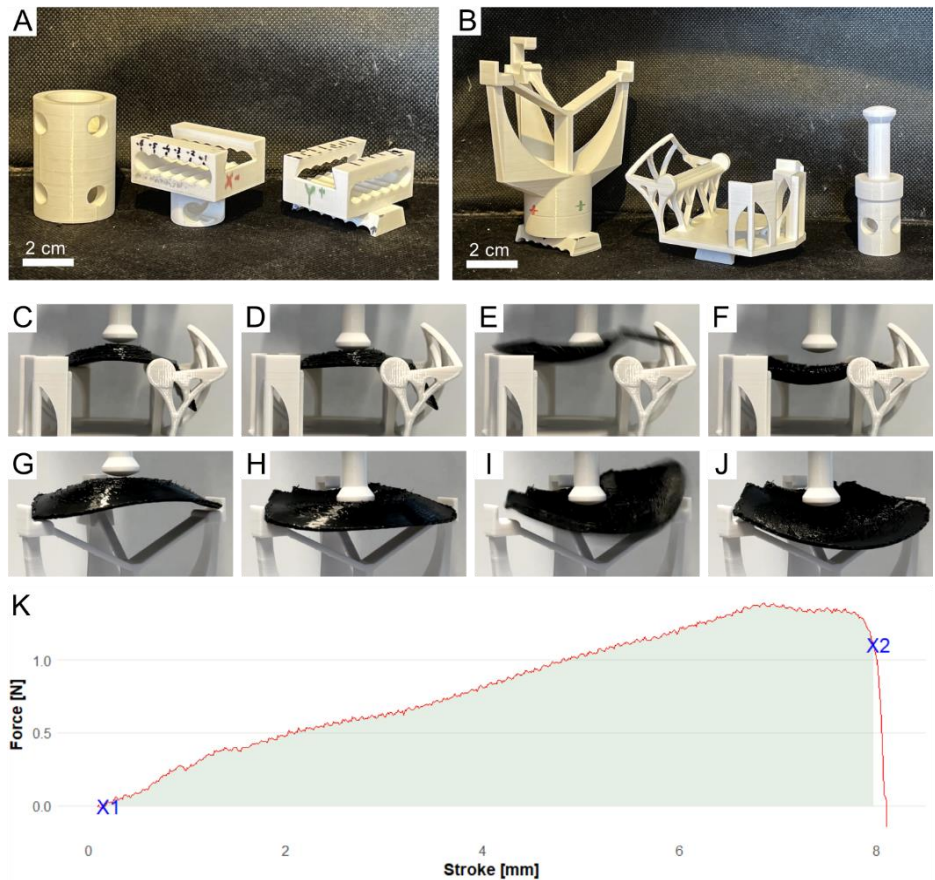

**Fig. 3.** Overview of the parts used for the setup in the universal testing machine (**A**, **B**). The connectors (A) with one of the two holders (B) on top form a multi-piece support structure that allows for adjustments in two directions and turning in increments of 30°. In B, the left structure is used for the ATL models, the middle for the SG models. On the right side is the indenter. Subfigures **C**–**F** show a motion sequence of an SG model snapping back from the loaded state, **G**–**J** show a motion sequence for an ATL model snapping. **K** shows an example graph of an ATL model snapping close. The loading force [N] is plotted against the machine's stroke [mm]. The work required is calculated as the integral from start of deformation at point X1 to snapping at point X2.

Both maximum force and required work for each model were consistently higher in the loading direction than in the snapping direction (Fig. 4). The multi-factorial ANOVAs showed significant differences between tested actuators and the direction of testing, for both maximum force (multi-factorial ANOVA, actuator × direction: df = 3, F = 471.5, $p \ll 0.005$) and work (multi-factorial ANOVA, actuator × direction: df = 3, F = 338.3, $p \ll 0.005$). The post-hoc tests revealed significant differences between all pairs, except between the maximum force measurements for snapping of the tapered ATL model and of the constant-thickness SG model (Tukey HSD, p = 0.96), and for the required work for snapping of the tapered SG model and the constant-thickness SG model (Tukey HSD, p = 0.99).

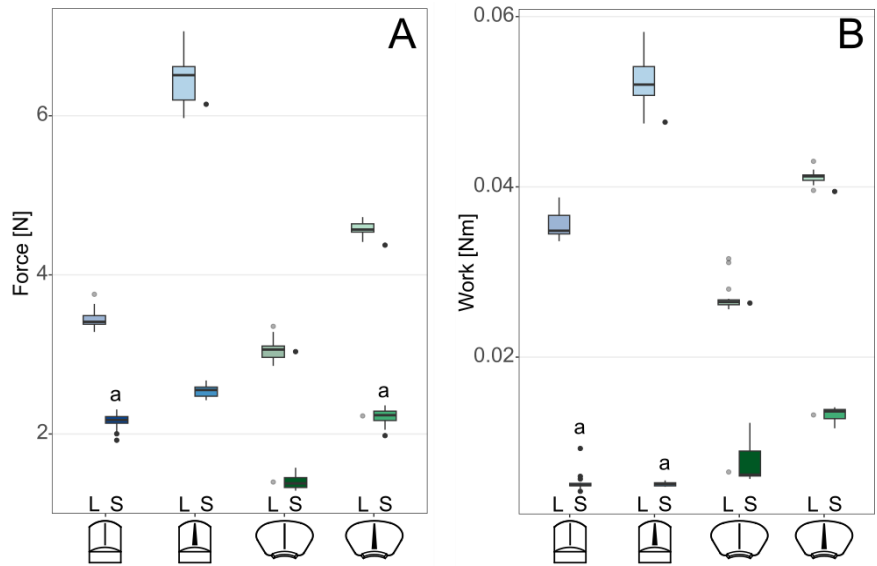

**Fig. 4.** Maximum force (**A**) and total work required (**B**) to change the state of the models. The icons below the plots indicate model (SG or ATL) and thickness variant (constant **I** or tapered ▲). "L" and "S" show the direction (Loading or Snapping). Within both plots, all combinations are significantly different (p < 0.005), except the boxplots marked with letters.

## 5    Discussion

Within this study, we investigated how different design parameters impact the bistable behavior of our Venus flytrap lobe actuator models, focusing on the impact of thickness, curvature, and tapering on the stability and needed actuation force of the lobes. The observed differences highlight that the ATL models with constant thickness and ellipsoidal curvature demonstrate higher stability in the loaded state and reduced force requirements for actuation compared to all other models. All four actuator models exhibited bistable behavior, switching between the initial and loaded state in a multi-step motion and snapping back with a sudden release of energy. Upon snapping (Fig. 3E and I), the models exhibited displacement, where they moved or shifted position, likely due to the release of stored energy. Only the tapered SG model in loading direction did not exhibit bistable behavior until the point of indentation was shifted 3 mm towards the hinge. This indicates that the ideal loading point needs to be considered for actuation of the opening motion in setting and resetting of a future gripper. From the results it is visible that the loading and snapping force in comparison to the other models is significantly higher in the tapered SG model. Thus, this indicates that the tapered SG model is the least suited for a gripper, as it necessitated more energy input for closing and opening, even after shifting the loading indentation point. The more natural ATL geometry necessitated lower force and work for loading and snapping in comparison to the SG models. This indicates that the SG models are too simplified and that the more ellipsoidal wider geometry of the ATL, which achieves a more stable state, is better suited for a gripper. The higher stability in the loaded state would prevent preliminary snapping during gripper operation and self-triggering over time, which was seen in preliminary tests in the SG models. To note is that the energy required to load each model was across all models greater than the energy needed for snapping it back. This indicates that adding prestress by loading the models could benefit the snap closure of a future gripper system. The system would release the prestress and stored energy during the snap closure, eliminating the need for additional energy for closure beyond what is required for triggering. The difference in energy levels of the SG and ATL models is also evident in the time difference between loading and snapping. The biological trap requires several hours to load and only a fraction of a second to snap closed [3, 4, 6]. The ATL model showed similar behavior, with both the loading and snapping motions propagating in similar ways, with the snapping being much faster.

The maximum measured force and work were consistently higher in the loading direction than in the snapping direction. This indicates that the energy threshold required for transition to the loaded open state is higher than that needed to snap to the closed state. Comparing all force and work measurements the post hoc tests showed only two pairs of model types with non-significant differences when compared. In the tapered and constant-thickness SG models, the work required to snap is similar, while the tapered model requires a higher energy input for loading. This indicates that the tapering had no effect on the snapping, and a negative impact on the loading force and stability of the models. The models with constant thickness required lower maximum force and work for loading than the same models with a tapered thickness. This is in contrast to the natural tapered lobes of the Venus flytrap. As such, we expected the tapered models to be more stable than the non-tapered models. One reason for this difference could be the material and production technique used, another may be that the different layers of the biological trap are active themselves during closure and contribute differently to the movement [1]. The preliminary designs with similar thickness ratios as the natural plant exhibited no bistability. Therefore, the current thickness gradient used in the tapered models may still be too high to reach the optimum that the natural Venus flytrap has reached. However, our current manufacturing methods limit the subtlety of the gradient we can implement. For printing thin structures, our FFF-printers have a relatively low resolution in the z-axis (>0.1 mm) which, combined with the curvature of the models, poses a challenge for implementing a smooth thickness gradient. Higher resolution 3D printing techniques such as SLA or DLP printing could improve this, but currently few suitable flexible materials exist for these manufacturing techniques. Based on our preliminary results, we derive that a slenderer, broader, non-tapered ellipsoid design achieves a more stable loaded state and are more suitable for FFF printed grippers.

In order to improve our current designs and further investigate the design parameters, we will benefit from a more in-depth motion analysis, which is work in progress. Based on these analyses, we then can compute the motion efficiency of the designs. Our current method involved indentation on a stabilized lobe, held on a holder with contact points on the lobe itself. When applied as a gripper, the lobes need to be mobile and thus the method of actuation would be different. Here we plan to use highly efficient actuation systems like hydraulics or environmental energy driven systems to load the gripper and mechanical triggering by touching the object to trigger snap grasping. For a general analysis however, the indentation method gives a good basic insight in the bistability of the shape, providing a repeatable method that can be used for a variety of lobe actuators. As mentioned above in this method the position of force application needs to be considered for future developments to find the ideal snapping point for the specific application, e.g., different trigger requirements or low-energy grippers. The next step for our analytical study will be a high-speed video analysis that will provide understanding of the propagation of the buckling motion as well as the curvature change during buckling and will give insights on whether the partial buckling motion benefits or impedes the snapping threshold and/or speed. In terms of speed, Zeng et al. [25] highlighted that geometric parameters have no influence on the snapping speed, while noting that it may be due to their definition of the speed. Our successful development of artificial lobes presents the possibility to isolate the factor of geometry in a model lobe and will make it possible to test whether it influences the snapping speed or not.

The comparison to other AVF systems is difficult as our system currently represents the first step towards a more fully 3D printable design for a soft AVF bistable gripper made from flexible material. The only 3D printable AVF systems [16, 18] lack the bistability of our system achieved through our geometry and need to be continuously actuated. Other bistable AVFs need high forces through pneumatics of 3.96 N to 7.35 N for triggering [19] or necessitate high electrical currents of above 4 kV [14, 20], and/or complex actuation systems to achieve bistability and to trigger fast snap closure [14, 16, 20]. So, in comparison we developed the first 3D printable bistable soft lobe requiring less force to load and trigger then other bistable AVF systems. Additionally, other AVF grippers lack the biaxial bending highlighted in this work. With our lobes, we aim to closely mimic the Venus flytrap's ability to rapidly transition between stable states, and so create a gripper that can perform fast, precise movements while retaining the adaptability and gentleness of a soft robotic system.

**Acknowledgments.** Research funded by the Deutsche Forschungsgemeinschaft (DFG, German Research Foundation) under Germany's Excellence Strategy – EXC-2193/1 – 390951807. RS acknowledges funding by the Klaus Tschira Boost Fund, a joint initiative of the German Scholars Organization and the Klaus Tschira Stiftung and the Freiburg Rising Stars Academy.